# Dynamic Stochastic Orienteering Problems
# for Risk-Aware Applications[*]

**Hoong Chuin Lau, William Yeoh, Pradeep Varakantham, Duc Thien Nguyen, Huaxing Chen**
Living Analytics Research Centre
Singapore Management University
Singapore 679668
{hclau,williamyeoh,pradeepv,dtnguyen,hxchen}@smu.edu.sg

## Abstract

Orienteering problems (OPs) are a variant of the well-known prize-collecting traveling salesman problem, where the salesman needs to choose a subset of cities to visit within a given deadline. OPs and their extensions with stochastic travel times (SOPs) have been used to model vehicle routing problems and tourist trip design problems. However, they suffer from two limitations – travel times between cities are assumed to be time independent and the route provided is independent of the risk preference (with respect to violating the deadline) of the user. To address these issues, we make the following contributions: We introduce (1) a dynamic SOP (DSOP) model, which is an extension of SOPs with dynamic (time-dependent) travel times; (2) a risk-sensitive criterion to allow for different risk preferences; and (3) a local search algorithm to solve DSOPs with this risk-sensitive criterion. We evaluated our algorithms on a real-world dataset for a theme park navigation problem as well as synthetic datasets employed in the literature.

## 1 Introduction

An orienteering problem (OP) is a planning problem where the goal is to find a sequence of vertices in a graph that maximizes the sum of rewards from those vertices subject to the constraint that the sum of edge lengths along that sequence is no larger than a threshold [34]. It was motivated by competitive orienteering sports where competitors start and end at specified locations and try to accumulate as much reward as possible from visiting checkpoints within a given deadline. Aside from this problem, researchers have also used OPs to model other problems like vehicle routing [11], production scheduling [2] and, more recently, tourist trip design problems [35].

One of the limitations of OPs is that it assumes that edge lengths are deterministic, which can be an inaccurate assumption in applications with uncertainty. Using vehicle routing problems as an example, the travel time between cities, which are modeled with edge lengths, depends on road congestion and is thus not deterministic. Thus, researchers have extended OPs to Stochastic OPs (SOPs), where edge lengths are now random variables that follow a given distribution and the goal is now to find a sequence that maximizes the sum of expected utilities from vertices in the sequence [6].

Although SOPs more accurately model applications with uncertainty, there are two assumptions that limit its applicability. Firstly, the assumption that the distribution of the random variables is time independent can still be inaccurate. Using the vehicle routing problem again as an example, the level of congestion of a road can change throughout the day – roads are more likely to be congested during commuting hours and less likely to be congested in the middle of the night. Thus, the distribution of travel times on the road should also change throughout the day. Therefore, in this paper, we extend SOPs to Dynamic SOPs (DSOPs).

Secondly, the expected utilities do not capture the risk preference of the user (with respect to violating the deadline) who will be employing the solution. In other words, the solution returned for a user with a hard deadline is the same as the one for a user with a soft deadline. Intuitively, the solution for the user with a soft deadline should be longer, and thus more rewarding, but riskier than the solution for the user with a hard deadline. To address this issue, we intro-duce a risk-sensitive criterion that captures this risk preference. To illustrate the applicability of DSOPs, we evaluated our algorithms on a real-world dataset for a theme park navigation problem as well as synthetic datasets employed in the literature.

---

[*]This research is supported by the Singapore National Research Foundation under its International Research Centre @ Singapore Funding Initiative and administered by the IDM Programme Office. We also thank Ajay S. Aravamudhan for his help with our experimental evaluations.

## 2 Motivating Domain

Ubiquitous computing has made substantial progress in recent years, particularly fueled by the increased prevalence of "smart" devices like smart phones. The widespread use of such devices presents a unique opportunity for the delivery of real-time contextualized and personalized information to users. For instance, operators of theme parks have now deployed smart phone applications that allow users to access real-time information like current queueing times at various attractions so that they can better plan their itinerary in the park.[1] Unfortunately, access to current queueing times only allows users to plan myopically, that is, to choose which attraction to go to next. If users also had access to predicted future queueing times, which park operators should be able to provide based on historical data, then the users might be able to make longer term plans that take that information into account. Our research is motivated by this exact problem, where a visitor to a theme park likes to choose (or be recommended) a sequence of attractions that optimizes his/her visitor experience subject to the constraint that the total travel and queueing time in the park is no larger than a threshold. In this paper, we conduct experiments on a real-world dataset for this problem.

## 3 Background

We now provide a brief description of OPs and SOPs.

### 3.1 OPs

The *orienteering problem* (OP) [34] is defined by a tuple $\langle G, T, R, v_1, v_n, H \rangle$. $G = \langle V, E \rangle$ is a graph with sets of vertices $V$ and edges $E$; $T : v_i \times v_j \to \mathbb{R}^+ \cup \{0, \infty\}$ specifies a finite non-negative travel time between vertices $v_i$ and $v_j$ if $e_{ij} \in E$ and $\infty$ otherwise; and $R : v_i \to \mathbb{R}^+ \cup \{0\}$ specifies a finite non-negative reward for each vertex $v_i \in V$. A solution in an OP is a Hamiltonian path over a subset of vertices including start vertex $v_1$ and end vertex $v_n$ and whose total travel time is no larger than $H$. Solving OPs optimally means finding a solution that maximizes the sum of rewards of vertices in its path. Researchers have shown that solving OPs optimally is NP-hard [11]. In this paper, we assume that the end vertex can be any arbitrary vertex.

---

[1]http://disneyparksmobile.com/ is one such example.

A simplified version of our motivating theme park navigation problem, where travel and queueing times are deterministic and static, can be modeled as an OP: $v_1$ corresponds to the entrance of the park, $v_n$ corresponds to the exit of the park, other vertices $v_i$ correspond to attractions in the park, and travel times $T(v_i, v_j)$ correspond to the sum of the travel time between attractions $v_i$ and $v_j$ and the queueing time at attraction $v_j$.

The start and end vertices in OPs are typically distinct vertices. In the special case where they are the same vertex, the problem is called a *orienteering tour problem* (OTP) [27]. The difference between both formulations is small. It is always possible to add a dummy edge with zero travel time between the start and end vertices to convert an OP to an OTP.

Researchers have proposed several exact branch-and-bound methods to solve OPs [18] including optimizations with cutting plane methods [20, 9]. However, since OPs are NP-hard, exact algorithms often suffer from scalability issues. Thus, constant-factor approximation algorithms [4] are necessary for scalability. Researchers also proposed a wide variety of heuristics to address this issue including sampling-based algorithms [34], local search algorithms [11, 7], neural network-based algorithms [38] and genetic algorithms [32]. More recently, Schilde et al. developed an ant colony optimization algorithm to solve a bi-objective variant of OPs [29].

### 3.2 Stochastic OPs

The assumption of deterministic travel times is unrealistic in many real-world settings. Using our motivating theme park navigation problem as an example, the travel time of a patron depends on factors like fatigue. Thus, researchers have extended OPs to *Stochastic OPs* (SOPs) [6], where travel times are now random variables that follow a given distribution, and the goal is to find a path that maximizes the sum of expected utilities from vertices in the path. The random variables are assumed to be independent of each other. The expected utility of a vertex is the difference between the expected reward and expected penalty of the vertex. The expected reward (or penalty) of a vertex is the reward (or penalty) of the vertex times the probability that the travel time along the path thus far is no larger (or larger) than $H$. More formally, the expected utility $U(v_i)$ of a vertex $v_i$ is

$$U(v_i) = P(a_i \leq H) R(v_i) - P(a_i > H) C(v_i)$$

where the random variable $a_i$ is the arrival time at vertex $v_i$ (that is, the travel time from $v_1$ to $v_i$), $R(v_i)$

is the reward of arriving at vertex $v_i$ before or at $H$ and $C(v_i)$ is the penalty of arriving at vertex $v_i$ after $H$. Campbell et al. have extended OP algorithms to solve SOPs including an exact branch-and-bound method and a local search method based on variable neighborhood search [6]. Gupta et al. introduced a constant-factor approximation algorithm for a special case of SOPs, where there is no penalty for arriving at a vertex after $H$ [12].

## 4 Dynamic Stochastic OPs

Stochastic OPs (SOPs) assume independence of travel time distributions across different edges. However, in many problems, there is a considerable dependence of travel times on the arrival time at a vertex. Using our motivating theme park navigation problem again as an example, the travel time of a user depends on factors like fatigue, and the level of fatigue of a patron increases as they spend more time in the park.

To capture dependencies between travel time distributions, we introduce an extension to SOPs called *Dynamic SOPs* (DSOPs). The key difference from SOPs is that the travel time distribution in DSOPs for moving from vertex $v_i$ to $v_j$ is dependent on the arrival time $a_i$ at vertex $v_i$. In this paper, we will assume $T_{i,j}$ to be a discrete set of distributions, where each element of the set corresponds to a range of values for $a_i$. Therefore, the travel time distribution for an arrival time of $a_i$ is represented as $T_{i,j}^{a_i}$ and hence the probability that travel time is $m$ is given by $T_{i,j}^{a_i}(m)$.

### 4.1 Risk-Sensitive Criterion

While expected utility is a good metric in general, the approach by [6] suffers from many limitations. Firstly, it is a point estimate solution which does not consider the "risk" profile of the user. By "risk", we do not refer to the term used in a financial sense, but rather the level of conservativeness measured in terms of the probability of completing the path within the deadline. In other words, a risk-seeking user will be prepared to choose a sequence of attractions that have a large utility, but with a higher probability of not completing the path within the deadline, compared to a risk-averse user who might choose a more "relaxed" path with lower utility. Secondly, the underlying measurement of expected utility is not intuitive in the sense that a utility value accrued at each attraction does not usually depend on the probability that the user arrives at the attraction by a certain time; but rather, the utility is accrued when the attraction is visited, and the user is concerned with visiting all the attractions (i.e. sum of utilities) within a certain time threshold.

Given the above consideration, we are interested in the problem that allows the user to tradeoff between the level of conservativeness (or risk) against the total utility. More precisely, given a value $0 \leq \epsilon \leq 1$, we are interested in obtaining a path, where the probability of completing the entire path within a deadline $H$ is at least $1 - \epsilon$. Or more precisely,

$$P(a_n \leq H) \geq 1 - \epsilon \qquad (1)$$

where $a_n$ is the arrival time at the last vertex of the path. The objective value is therefore inversely proportional to the value of $\epsilon$.

## 5 Completion Probability Approximations

In this section, we describe two ways of approximating the completion probability $P(a_n \leq H)$, which is used in Equation 1. Given the order $\pi = \langle v_1, v_2, \ldots, v_k, v_n \rangle$, we can use the following expression to compute the completion probability:

$$\begin{aligned}
P(a_n \leq H) = & \\
& \int_{a_n=0}^{a_n \leq H} \int_{a_k=0}^{a_k=a_n} \int_{a_{k-1}=0}^{a_{k-1}=a_k} \cdots \int_{a_1=0}^{a_1=a_2} \\
& T_{k,n}^{a_k}(a_n - a_k) T_{k-1,k}^{a_{k-1}}(a_k - a_{k-1}) \cdots T_{1,2}^{a_1}(a_2 - a_1) \\
& d(a_1) \, d(a_2) \ldots d(a_k) \, d(a_n) \qquad (2)
\end{aligned}$$

where $a_n$ is the arrival time at the exit node, and we capture the dependencies on arrival times at each of the vertices by reducing the range of feasible arrival times (for the integrals) based on the previous activities in the order of vertices. Unfortunately, the computation of the expression is expensive since the integrals have to be computed sequentially. To provide an intuition for the time complexity, computing triple integrals take around 30 minutes with an exponential distribution (most scalable of all distributions with integration) on our machine using the Matlab software. To address this issue of scalability, we introduce two approximation approaches, a sampling-based approach and a matrix-based approach.

### 5.1 Sampling-based Approach

One can approximate the completion probability $P(a_n \leq H)$ of a path by randomly sampling the travel time distributions for each edge along the path, and checking if the arrival time $a_n$ at the last vertex exceeds $H$. For example, assume that we want to compute $P(a_n \leq H)$ for the path $\pi = \langle v_1, v_2, \ldots, v_k, v_n \rangle$. Using the starting time $a_1$, we generate a travel time sample from the distribution $T_{1,2}^{a_1}$ to represent the travel time from vertex $v_1$ to vertex $v_2$, which is also the arrival time $a_2$ at vertex $v_2$. We then generate a travel time sample from the distribution $T_{2,3}^{a_2}$ to represent the travel time from vertex $v_2$ to vertex $v_3$. The arrival time $a_3$ at vertex $v_3$ is thus the sum of both travel times. We continue this process until we generate a travel time sample to represent the travel time from vertex $v_k$ to vertex $v_n$, and the arrival time $a_n$ is thus the sum of all travel times. We count this entire process as a single sample. We can then approximate

$$P(a_n \leq H) \approx \hat{P}(a_n \leq H) = \frac{N^+}{N} \quad (3)$$

where $N^+$ is the number of samples whose arrival time $a_n \leq H$ is no larger than the deadline $H$ and $N$ is the total number of samples. Unfortunately, this approach does not provide any theoretical guarantees on whether Equation 1 is truly satisfied. However, as we increase the number of samples, the approximation for the actual distribution becomes tighter.

### 5.2 Matrix-based Approach

Alternatively, one can exploit the fact that the dependencies are primarily due to arrival time at a vertex and not on the entire order of vertices before the current vertex. At a higher level, it implies that the underlying problem is Markovian and hence we can decompose the expression of Equation 2. We also make conservative estimates of the probability such that we can provide theoretical guarantees on whether Equation 1 is truly satisfied.

The key ideas here are (1) to divide the possible arrival times $a_i$ at vertex $v_i$ into a finite number of ranges $\mathbf{r}_{i,1}, \mathbf{r}_{i,2}, \ldots, \mathbf{r}_{i,k}$, where $\mathbf{r}_{i,j}$ is the $j$-th range of arrival time at vertex $v_i$ and (2) to pre-compute for all pairs of vertices $v_i$ and $v_j$ a conservative estimate $\hat{P}(a_j \in \mathbf{r}_{j,q}|a_i \in \mathbf{r}_{i,p})$ of the probability $P(a_j \in \mathbf{r}_{j,q}|a_i \in \mathbf{r}_{i,p})$ of transitioning between ranges of arrival times $\mathbf{r}_{i,p}$ and $\mathbf{r}_{j,q}$. Thus, we can now decompose the expression of Equation 2 to an expression that exploits the Markovian property along with ranges of arrival times:

$$P(a_n \leq H) =$$
$$\sum_i P(a_1 \in \mathbf{r}_{1,i}) \cdot \sum_j P(a_2 \in \mathbf{r}_{2,j}|a_1 \in \mathbf{r}_{1,i})$$
$$\cdots \sum_y P(a_k \in \mathbf{r}_{k,y}|a_{k-1} \in \mathbf{r}_{k-1,x})$$
$$\cdot \sum_z P(a_n \in \mathbf{r}_{n,z}|a_k \in \mathbf{r}_{k,y})$$

and conservatively approximate it by

$$\hat{P}(a_n \leq H) =$$
$$\sum_i \hat{P}(a_1 \in \mathbf{r}_{1,i}) \cdot \sum_j \hat{P}(a_2 \in \mathbf{r}_{2,j}|a_1 \in \mathbf{r}_{1,i})$$
$$\cdots \sum_y \hat{P}(a_k \in \mathbf{r}_{k,y}|a_{k-1} \in \mathbf{r}_{k-1,x})$$
$$\cdot \sum_z \hat{P}(a_n \in \mathbf{r}_{n,z}|a_k \in \mathbf{r}_{k,y})$$

It is clear that $\hat{P}(a_n \leq H) \leq P(a_n \leq H)$ is a conservative estimate if $\hat{P}(a_j \in \mathbf{r}_{j,q}|a_i \in \mathbf{r}_{i,p}) \leq P(a_j \in \mathbf{r}_{j,q}|a_i \in \mathbf{r}_{i,p})$ are all conservative estimates. $\hat{P}(a_1 \in \mathbf{r}_{1,i})$ depends on the starting time at vertex $v_1$, which is provided as an input. We now describe how to compute the other probabilities $\hat{P}(a_j \in \mathbf{r}_{j,q}|a_i \in \mathbf{r}_{i,p})$. If the range $\mathbf{r}_{i,p}$ contains only a single point $a_i$, then

$$P(a_j \in \mathbf{r}_{j,q}|a_i \in \mathbf{r}_{i,p}) = P(a_j \in \mathbf{r}_{j,q}|a_i)$$
$$= \int_{a_j \in \mathbf{r}_{j,q}} T_{i,j}^{a_i}(a_j - a_i) \, d(a_j)$$
$$(4)$$

However, the realization of the random variable $a_i$ only occurs at runtime, and computing the integral in Equation 4 at runtime is expensive. Thus, we would like to compute a conservative estimate $\hat{P}(a_j \in \mathbf{r}_{j,q}|a_i \in \mathbf{r}_{i,p})$ of probability $P(a_j \in \mathbf{r}_{j,q}|a_i \in \mathbf{r}_{i,p})$ for all possible realizations of $a_i \in \mathbf{r}_{i,p}$. We thus compute them as follows:

$$\hat{P}(a_j \in \mathbf{r}_{j,q}|a_i \in \mathbf{r}_{i,p}) =$$
$$\min_{a_i \in \mathbf{r}_{i,p}} \int_{a_j \in \mathbf{r}_{j,q}} T_{i,j}^{a_i}(a_j - a_i) \, d(a_j) \quad (5)$$

The value in the integral is the probability of the arrival time $a_j$ to be in the range $\mathbf{r}_{j,q}$ for a given value of $a_i$. Thus, by taking the minimum of these probabilities over all possible values of $a_i$ in the range $\mathbf{r}_{i,p}$, the conditional probability $\hat{P}(a_j \in \mathbf{r}_{j,q}|a_i \in \mathbf{r}_{i,p}) \leq P(a_j \in \mathbf{r}_{j,q}|a_i \in \mathbf{r}_{i,p})$ is a conservative estimate of the true probability.

Once all the probabilities are pre-computed, they form transition matrices

$$\mathbf{P}_{i,j} =$$
$$\begin{pmatrix} \hat{P}(a_j \in \mathbf{r}_{j,1})|a_i \in \mathbf{r}_{i,1} & \hat{P}(a_j \in \mathbf{r}_{j,2})|a_i \in \mathbf{r}_{i,1} & \cdots \\ \hat{P}(a_j \in \mathbf{r}_{j,1})|a_i \in \mathbf{r}_{i,2} & \hat{P}(a_j \in \mathbf{r}_{j,2})|a_i \in \mathbf{r}_{i,2} & \cdots \\ \cdots & \cdots & \cdots \end{pmatrix}$$
$$(6)$$

which represent the transition probabilities from vertices $v_i$ to $v_j$. Finally, to compute $\hat{P}(a_n \leq H)$, we compute the multiplication of matrices $\mathbf{P}_1 \cdot \mathbf{P}_{1,2} \cdot \mathbf{P}_{2,3} \cdots \mathbf{P}_{k-1,k} \cdot \mathbf{P}_{k,n}$ and in the resultant matrix, sum up all the probabilities for ranges of arrival times $a_n$ that are less than or equal to the deadline $H$.

**Algorithm 1:** Local Search Algorithm

```
   /* Generate Initial Solution              */
 1 currentPath = ConstructionHeuristic()
   /* Make Local Improvements                */
 2 bestPath = currentPath
 3 numIterNoImprove = 0
 4 currentMetric = random metric
 5 T = starting temperature
 6 for iterations = 1 to maxIterations do
 7    T = T · ΔT
 8    Z = numIterNoImprove / (2·maxIterNoImprove)
      /* Perform 2-Opt Operation on currentPath */
 9    currentPath = 2-Opt(currentPath)
      /* Remove Vertices from currentPath     */
10    while currentPath is infeasible OR rand() ≤ Z do
11       remove the second last vertex from currentPath
12    end
      /* Insert Vertices to currentPath       */
13    neighborPath = Insert(currentPath, currentMetric)
      /* Update currentPath and bestPath      */
14    ΔR = neighborPath.reward − currentPath.reward
15    if ΔR > 0 OR rand() ≤ e^{ΔR/T} then
16       currentPath = neighborPath
17    end
18    if currentPath.reward > bestPath.reward then
19       bestPath = currentPath
20       numIterNoImprove = 0
21    else
22       numIterNoImprove = numIterNoImprove + 1
23       if numIterNoImprove > maxIterNoImprove then
24          currentMetric = new random metric
25          numIterNoImprove = 0
26       end
27    end
28 end
29 return bestPath
```

## 6 DSOP Algorithms

In this section, we describe a branch-and-bound algorithm and a local search algorithm that solves DSOPs.

### 6.1 Branch-and-Bound Algorithm

We provide a depth-first branch-and-bound algorithm, where the root of the search tree is the start vertex and the children of a vertex are all the unvisited vertices minus the exit vertex. The branch of an arbitrary vertex thus represents the path from the start vertex to that vertex. The value of a vertex is the sum of rewards of all vertices along its branch. The algorithm prunes the subtree of a vertex if it fails to satisfy our risk-sensitive criterion. For example, assume that a vertex $v_k$ is on the branch $\pi = \langle v_1, v_2, \ldots, v_k \rangle$, where vertex $v_i$ is on the $i$-th position on the branch. The algorithm prunes the subtree rooted at vertex $v_k$ if the condition in Equation 1 is not satisfied if one appends the exit vertex to the end of the path. The algorithm returns the vertex with the largest value and the branch of that vertex with the exit vertex appended at the end of the path as the best solution that satisfies the risk-sensitive criterion.

### 6.2 Local Search Algorithm

Unfortunately, the branch-and-bound algorithm suffers from scalability issues as the size of the search tree is exponential in the number of vertices in the graph. We thus introduce a local search algorithm that is based on the standard two-phase approach – a construction heuristic to generate an initial solution followed by local improvements on that solution.

#### 6.2.1 Construction Heuristic

The construction heuristic is a greedy insertion algorithm that greedily inserts the best unvisited vertex at the best position in the current path according to a given metric. The algorithm begins with the path that starts at the start vertex and immediately exits at exit vertex, and it terminates when it can no longer insert any attraction at any position without violating the condition in Equation 1.

In this paper, we use the following metric to evaluate the value of inserting vertex $v_i$ at position $p$: $\frac{\Delta R}{1+\Delta P}$, where $\Delta R$ and $\Delta P$ is the gain in reward and probability, respectively, for inserting vertex $v_i$ at position $p$. Thus, $\Delta R = R(v_i)$, which is the reward of vertex $v_i$, and $\Delta P = \hat{P}(a_n \leq H) - \hat{P}'(a_n \leq H)$, where $\hat{P}'(a_n \leq H)$ and $\hat{P}(a_n \leq H)$ is the probability of arriving at the exit vertex before and after insertion, respectively. We use the same approach of multiplying transition matrices described in Section 6.1 to speed up the computation of the probabilities. Finally, we add 1 to the gain in probabilities such that the denominator is greater than 0.

This metric is motivated by similar metrics in knapsack problems, namely the utility of an item is the ratio between the reward and size of that item [23]. We have also tried 4 other variants of the above metric, namely (1) $\frac{1}{1+\Delta P}$, (2) $\Delta R$, (3) $\frac{(\Delta R)^2}{1+\Delta P}$, (4) $\frac{\Delta R}{\sqrt{1+\Delta P}}$, where we ignored the effects of rewards in (1) and probabilities in (2), and we amplified the effects of rewards in (3) and probabilities in (4). However, our chosen metric was shown to outperform these 4 variants empirically.

#### 6.2.2 Local Improvements

We use a hybrid approach that consists of a variable neighborhood search combined with simulated annealing to locally improve our initial solution found by the construction heuristic. Algorithm 1 shows the pseudocode of this algorithm. After constructing the ini-tial solution (line 1), the algorithm iteratively runs the following four phases until the maximum number of iterations is reached (line 6):

**Phase 1:** If the path contains at least two vertices (not including the start and end vertices), then the algorithm performs a 2-Opt operation, that is, it randomly swaps two of these vertices (line 9).

**Phase 2:** If the path is not feasible, that is, it does not satisfy Equation 1, then the algorithm repeatedly removes the second last vertex until the path is feasible. (The algorithm does not remove the last vertex because it is the exit vertex.) Once the path is feasible, the algorithm repeatedly removes the second last vertex probabilistically (lines 10-12).[2]

**Phase 3:** The algorithm repeatedly inserts unvisited vertices greedily similar to the construction heuristic (line 13). The difference here is that the metric used can be one of five different metrics, either the metric chosen for the construction heuristics or one of its four variants described above. The algorithm starts by choosing one of the five metrics randomly (line 4). If there are no improvements in *maxIterNoImprove* iterations, the algorithm chooses a new different metric randomly (lines 25-26). These different metrics correspond to the different "neighborhoods" in our variable neighborhood search.

**Phase 4:** The algorithm then updates the current path to the new neighboring path, which is a result from inserting unvisited vertices in Phase 3, if the new path is a better path or with a probability that depends on the simulated annealing temperature (lines 14-17).

**Reusing Matrix Computations:** We re-compute the completion probability $\hat{P}(a_n \leq H)$ whenever we make a local move during the search, that is, when (a) two vertices are swapped, (b) a vertex is removed, and/or (c) a vertex is inserted. To compute these probabilities efficiently, we store the results of the products of transition matrices for subsets of vertices. For example, in a path $\pi = \langle v_1, v_2, \ldots, v_i, \ldots, v_j, \ldots \rangle$, if we swap vertices $v_i$ and $v_j$, then the product of transition matrices for the vertices before $v_i$, the product of matrices for the vertices between $v_{i+1}$ and $v_{j-1}$, and the product of matrices for the vertices after $v_{j+1}$ remain unchanged. By storing all of these intermediate results, it is possible to make the computation of probabilities very efficient. However, it requires a significant amount memory for larger problems. In this paper, we store only the products of matrices for the vertices between the start vertex and every subsequent vertex in the path except for the exit vertex. For example, for a path $\pi = \langle v_1, v_2, v_3, v_n \rangle$, we store the product of matrices for vertices $v_1$ and $v_2$, which is $\hat{P}(a_2 \leq H)$, and the product of matrices for vertices $v_1$, $v_2$ and $v_3$, which is $\hat{P}(a_3 \leq H)$. While this is not the most efficient approach, it provides a good tradeoff between memory requirement and efficiency.

## 7 Experimental Results

We now empirically demonstrate the scalability of our approaches on synthetic datasets employed in the literature as well as a real-world dataset for a theme park navigation problem. We run our experiments on a 3.2GHz Intel i5 dual-core CPU with 12GB memory, and we set the parameters for the local search algorithm as follows: we set *maxIterNoImprove* to 50, *maxIterations* to 1500, $T$ to 0.1 and $\Delta T$ to 0.99. We divide each travel time distribution to 100 ranges for the matrix-based computations and use 1000 samples for the sampling-based computations.

### 7.1 Synthetic Dataset Results

Our synthetic dataset is based on the dataset provided in [34] with 32 vertices. We assume that the total travel time of each edge is the sum of the travel time between the two vertices connected by that edge and the queueing time at the target vertex of that edge. As in [6], we assume that the total travel time of each edge is a gamma distribution, whose mean is the Euclidean distance between the vertices connected by that edge. We vary the shape parameter $2 \leq k \leq 9$ and scale parameter $1 \leq \theta \leq 4$ such that the mean of the values $\mu \approx k\theta$ is approximately equal to the product of the shape parameter $k$ and the scale parameter $\theta$. A gamma distribution with $k = 1$ is an exponential distribution. As we wanted a more normal-like distribution, we did not include this value of $k$ in our experiments. We also bound the possible values of $k$ such that shape of the distributions across time ranges do not vary significantly, and we use the same bound on the possible values of $\theta$ as in [6]. Lastly, we set rewards for each vertex to a random number between 1 and 100.

Table 1 shows our results for the construction heuristic algorithm (labeled CH) and local search algorithm (labeled LS), where we calculate the completion probability of a path (see Equation 1) using both the matrix-based approach (see Equations 5 and 6) and the sampling-based approach (see Equation 3). We report the completion probability of the best path found by the local search algorithm using the matrix-based approach (labeled $P_M$) and the sampling-based approach (labeled $P_S$). We also report the the percentage of improvement in the reward of the path found by the local search algorithm compared to the path found by the construction heuristic algorithm (denoted

---
[2]The rand() function returns a random number in [0,1].

(a) Results averaged across all deadlines $H$ and risk parameters $\epsilon$

|  | Matrix-based Approach ||||  Sampling-based Approach ||||
|  | Rewards || Runtimes (s) || Rewards || Runtimes (s) ||
|  | CH | LS | CH | LS | CH | LS | CH | LS |
|---|---|---|---|---|---|---|---|---|
| $\theta = 1$ | 87  | 88  (0.50) | 0.5 | 568 | 876 | 1033 (18.75) | 5.3 | 2443 |
| $\theta = 2$ | 129 | 134 (1.65) | 0.8 | 987 | 695 | 792  (17.03) | 2.7 | 1477 |
| $\theta = 3$ | 123 | 133 (3.97) | 0.8 | 904 | 533 | 569  (6.63)  | 1.3 | 716  |
| $\theta = 4$ | 140 | 140 (0.00) | 0.8 | 864 | 406 | 428  (6.39)  | 0.7 | 388  |

(b) Results averaged across all scale parameters $\theta$ and risk parameters $\epsilon$

|  | Matrix-based Approach ||||  Sampling-based Approach ||||
|  | Rewards || Runtimes (s) || Rewards || Runtimes (s) ||
|  | CH | LS | CH | LS | CH | LS | CH | LS |
|---|---|---|---|---|---|---|---|---|
| $H = 20$  | 28  | 28  (0.00) | 0.2 | 238  | 193  | 220  (12.71) | 0.2 | 221  |
| $H = 40$  | 94  | 94  (0.11) | 0.5 | 520  | 432  | 498  (15.00) | 0.8 | 690  |
| $H = 60$  | 138 | 141 (1.43) | 0.8 | 862  | 657  | 732  (11.10) | 2.0 | 1303 |
| $H = 80$  | 155 | 160 (2.00) | 0.9 | 1050 | 847  | 952  (11.35) | 3.7 | 1858 |
| $H = 100$ | 185 | 196 (4.12) | 1.3 | 1485 | 1008 | 1126 (10.84) | 5.7 | 2208 |

(c) Results averaged across all deadlines $H$ and scale parameters $\theta$

|  | Matrix-based Approach |||||| Sampling-based Approach ||||||
|  | Rewards || Runtimes (s) || $P_M$ | $P_S$ | Rewards || Runtimes (s) || $P_M$ | $P_S$ |
|  | CH | LS | CH | LS |  |  | CH | LS | CH | LS |  |  |
|---|---|---|---|---|---|---|---|---|---|---|---|---|
| $\epsilon = 0.1$ | 1   | 1   (0.00) | 0.1 | 168  | 1.00 | 1.00 | 507 | 605 (18.48) | 1.7 | 1077 | 0.17 | 0.90 |
| $\epsilon = 0.2$ | 46  | 46  (0.00) | 0.2 | 332  | 0.90 | 0.99 | 585 | 669 (13.98) | 2.1 | 1186 | 0.15 | 0.81 |
| $\epsilon = 0.3$ | 113 | 119 (3.38) | 0.6 | 768  | 0.79 | 0.99 | 643 | 711 (10.03) | 2.6 | 1270 | 0.13 | 0.73 |
| $\epsilon = 0.4$ | 194 | 197 (1.23) | 1.1 | 1248 | 0.66 | 0.97 | 679 | 757 (10.81) | 2.8 | 1376 | 0.09 | 0.63 |
| $\epsilon = 0.5$ | 246 | 256 (3.05) | 1.6 | 1640 | 0.54 | 0.95 | 725 | 785 (7.71)  | 3.3 | 1371 | 0.07 | 0.55 |

Table 1: Experimental Results for Simpler Synthetic Datasets

in parentheses beside the local search rewards). The branch-and-bound algorithm successfully terminated for problems with small deadlines $H$ and risk parameters $\epsilon$ only. Therefore, we did not tabulate its results as it is unfair to only consider successful runs in computing them. (We do not know the rewards and completion probabilities for unsuccessful runs.) We make the following observations:

- Table 1(a) shows that for the matrix-based approach, the solution rewards increase between $\theta = 1$ and $\theta = 2$, and remain relatively unchanged for larger values of $\theta$. As $\theta$ increases, the variance of the gamma distributions increases as well. When $\theta = 1$, only very few ranges have non-zero transition probabilities computed with Equation 5. As a result, adding an additional edge to a solution can result in a significant decrease in completion probability. With larger values of $\theta$, more ranges have non-zero transition probabilities, but the number of ranges and transition probabilities do not change much with increasing values of $\theta$. Thus, the path length and, consequently, reward and runtime, typically increases as $\theta$ increases from 1 to 2, but remains relatively unchanged for larger values of $\theta$. The runtime depends on the path length because the number of positions to check to find the best position to insert a vertex, which is done by the construction heuristic algorithm and phase 3 in the local improvement phase, depends on the path length.

  On the other hand, for the sampling-based approach, the solution rewards decrease as $\theta$ increases. Since the sampling probabilities are relatively accurate representations of the true probabilities, as the variance increases, adding an additional edge to a solution can result in a significant decrease in completion probability. Thus, the path length and, consequently, reward and runtime, typically decreases as $\theta$ increases.

- Table 1(a) also shows that as $\theta$ increases, for the sampling-based approach, the improvement of the local search algorithm over the construction heuristic algorithm decreases. The reason is that as the variance of the gamma distributions increases, there is less distinction between the different gamma distributions. Thus, many of the neighboring solutions are very similar to the solution found by the construction heuristic algorithm. For the matrix-based approach, the improvements are all negligible. The path lengths are short (with 1-3 vertices excluding the start and end vertices), and thus there is no much room

(a) Results averaged across all risk parameters $\epsilon$

|  | Matrix-based Approach |  |  |  | Sampling-based Approach |  |  |  |
|---|---|---|---|---|---|---|---|---|
|  | Rewards |  | Runtimes (s) |  | Rewards |  | Runtimes (s) |  |
|  | CH | LS | CH | LS | CH | LS | CH | LS |
| $H = 20$ | 29 | 29 (0.00) | 0.2 | 262 | 430 | 537 (30.35) | 0.9 | 1115 |
| $H = 40$ | 102 | 103 (0.43) | 0.5 | 571 | 864 | 1052 (22.27) | 3.8 | 4526 |
| $H = 60$ | 147 | 160 (6.77) | 0.7 | 915 | 1156 | 1394 (20.92) | 7.5 | 8255 |
| $H = 80$ | 181 | 183 (0.44) | 1.1 | 1326 | 1503 | 1588 (5.73) | 13.6 | 9339 |
| $H = 100$ | 247 | 263 (5.11) | 1.8 | 2004 | 1579 | 1644 (4.14) | 15.3 | 7716 |

(b) Results averaged across all deadlines $H$

|  | Matrix-based Approach |  |  |  |  |  | Sampling-based Approach |  |  |  |  |  |
|---|---|---|---|---|---|---|---|---|---|---|---|---|
|  | Rewards |  | Runtimes (s) |  | $P_M$ | $P_S$ | Rewards |  | Runtimes (s) |  | $P_M$ | $P_S$ |
|  | CH | LS | CH | LS |  |  | CH | LS | CH | LS |  |  |
| $\epsilon = 0.1$ | 16 | 16 (0.00) | 0.1 | 215 | 0.98 | 1.00 | 1004 | 1162 (24.68) | 6.9 | 6165 | 0.00 | 0.90 |
| $\epsilon = 0.2$ | 83 | 83 (0.00) | 0.4 | 500 | 0.87 | 0.97 | 1071 | 1222 (19.87) | 8.2 | 6237 | 0.00 | 0.81 |
| $\epsilon = 0.3$ | 141 | 144 (1.37) | 0.7 | 975 | 0.78 | 0.97 | 1109 | 1255 (18.38) | 8.1 | 6561 | 0.00 | 0.73 |
| $\epsilon = 0.4$ | 204 | 224 (9.17) | 1.2 | 1448 | 0.61 | 0.95 | 1144 | 1264 (11.25) | 8.5 | 6054 | 0.00 | 0.64 |
| $\epsilon = 0.5$ | 264 | 272 (2.23) | 1.8 | 1940 | 0.55 | 0.87 | 1205 | 1311 (9.24) | 9.6 | 5934 | 0.00 | 0.56 |

Table 2: Experimental Results for More Difficult Synthetic Datasets

|  | Rewards (Peak Days) |  |  |  |  | Rewards (Non-Peak Days) |  |  |  |  |
|---|---|---|---|---|---|---|---|---|---|---|
|  | $H = 2$ | $H = 4$ | $H = 6$ | $H = 8$ | $H = 10$ | $H = 2$ | $H = 4$ | $H = 6$ | $H = 8$ | $H = 10$ |
| $\epsilon = 0.1$ | 474 | 485 | 488 | 508 | 558 | 579 | 579 | 579 | 579 | 579 |
| $\epsilon = 0.2$ | 474 | 485 | 488 | 508 | 558 | 579 | 579 | 579 | 579 | 578 |
| $\epsilon = 0.3$ | 474 | 485 | 507 | 508 | 557 | 620 | 620 | 620 | 621 | 624 |
| $\epsilon = 0.4$ | 474 | 485 | 505 | 549 | 558 | 620 | 627 | 627 | 634 | 634 |
| $\epsilon = 0.5$ | 474 | 485 | 507 | 549 | 557 | 646 | 649 | 646 | 644 | 646 |

Table 3: Experimental Results for Real-World Theme Park Dataset

for improvement. We confirm this result with the branch-and-bound algorithm, where it found similar paths to those found by the local search algorithm for problems where it successfully terminated.

- Tables 1(b) and 1(c) show that as $H$ or $\epsilon$ increases, the solution reward increases for both matrix- and sampling-based approaches, which is to be expected. Similarly, the runtime also increases since the number of positions to check to find the best position to insert a vertex also increases.

- Table 1(c) shows that the completion probabilities $P_M$ and $P_S$ are all no less than $1 - \epsilon$ for the matrix- and sampling-based approaches, respectively, which is to be expected.

We observe that the problems in this dataset are relatively easy as all gamma distributions have the same scale parameter and their means satisfy the triangle inequality. Thus, we modified the dataset to increase its difficulties in the following ways: (a) we choose the scale parameter $\theta$ of the gamma distributions for each edge randomly between 1 and 4 such that not all edges have distributions with the same scale parameter, and (b) we change the shape parameter $k$ of the gamma distributions for some subset of edges such that their means no longer satisfy the triangle inequality. Table 2 shows our results for this more difficult synthetic dataset. We make the same observations here as in the simpler dataset with the exception that the improvements of the local search algorithm over the construction heuristic algorithm is now up to 30% as opposed to 18% earlier.

Overall, using the sampling-based approach, the local search algorithm provides reasonably better solutions compared to the construction heuristic algorithm. However, it is not guaranteed that these solutions are feasible, that is, they satisfy Equation 1. However, the feasibility likelihood increases with the number of samples. Thus, this approach is better suited for users without strict feasibility requirements. On the other hand, solution feasibility is guaranteed for algorithms using the matrix-based approach. Unfortunately, the local search algorithm fails to reasonably improve on the solutions found by the construction heuristic algorithm. Thus, the construction heuristic algorithm using the matrix-based approach is better suited for users with strict feasibility requirements.

### 7.2 Real-World Dataset Results

For our real-world theme park dataset, the total travel time of each edge is also a gamma distribution. We choose the scale parameter $\theta$ and shape parameter $k$ such $\mu \approx k\theta$ and $\sigma^2 \approx k\theta^2$, where $\mu$ and $\sigma^2$ is the mean and variance, respectively, of our data points (across several months) for the sum of travel time and queueing time. Similar to the synthetic datasets, we also set rewards for each vertex to a random number between 1 and 100. However, we do not bound the possible values of the scale and shape parameters $\theta$ and $k$ such that the gamma distributions approximate the data points as accurately as possible. Lastly, we segment our data points into two categories, peak days and non-peak days,[3] and present results for both categories.

Table 3 shows our results for the local search algorithm using the sampling-based approach to compute the completion probabilities; the deadline $H$ is measured in hours. We only show the solution rewards as the other trends are similar to those observed for the synthetic datasets. Similar to the synthetic datasets, as $H$ and $\epsilon$ increase, the solution reward also typically increases. Additionally, the solution rewards for non-peak days are larger than those for peak days. The reason is that the queueing time at an attraction is smaller on non-peak days than on peak days. Thus, it is possible to visit more attractions, and thus accrue more rewards, on non-peak days than on peak days.

## 8 Related Work

OPs have a long history and has been known by a variety of other names including selective TSPs [18], maximum collection problems [16] and bank robber problems [1]. Vansteenwegen et al. recently presented a broad overview of the problem, its variants and associated solution methods [36]. In contrast, to the best of our knowledge, there has not been much work in stochastic variants of OP thus far. Aside from the work on SOPs [6], two closely related problems with stochastic travel times are time-constrained TSPs with stochastic travel and service times [33] and stochastic selective TSPs [31]. These problems assume that the traveling time distributions are time independent, unlike our dynamic SOPs. Lastly, aside from travel time, researchers have also investigated stochasticity in the reward values of vertices [15].

SOPs also bear some similarity with Markov random fields (MRFs) [37] and Bayesian networks [28]. They are both graphical models, where nodes in a graph correspond to random variables and edges in a graph correspond to potential functions between pairs of random variables. While MRF graphs can be cyclic, Bayesian network graphs are strictly acyclic. The goal in these two models is to compute the maximum a posteriori (MAP) assignment, which is the most probable assignment to all the random variables of the underlying graph in MRFs [37, 17, 30] and Bayesian networks [24, 14, 39]. Thus, the main difference between MAP assignment problems and SOPs is that MAP assignment problems are *inference* problems while SOPs are *planning* problems. DSOPs can potentially be represented using the TiMDP model [5]. However, the objective in TiMDPs is to maximize expected reward, which is unlike in DSOPs, where we also consider the robustness criterion.

With respect to modeling and accounting for different risk preferences, there are generally the following three approaches: (1) Stochastic dominance, whose theory was developed in statistics and economics [19, 13]. Stochastic dominance defines partial orders on the space of random variables and allow for pairwise comparison of different random variables. (2) Mean-risk analysis, whose models originate from finance. They include the well known mean-variance optimization model in portfolio optimization, where the variance of the return is used as the risk functional [21]. (3) Chance constraints or percentile optimization, whose models were initiated and developed in operations research [22, 25]. Recently, researchers have provided a thorough overview of the state-of-the-art of the optimization theory with chance constraints [26]. Our approach of defining a risk-sensitive measure that allows the user to specify a level of risk (failure tolerance) is along the lines of using chance constraints to model and account for different risk preferences. While it has been applied to solve planning and scheduling problems [3, 8, 10], to the best of our knowledge, it has yet to be applied to solve OPs.

## 9 Conclusions

Researchers have used OPs to model vehicle routing and tourist trip design problems. However, OPs assume that the travel times are independent of the time of day and the route returned is independent of the risk preference of the user. Therefore, we make the following contributions in this paper: (1) we introduce a dynamic and stochastic OP (DSOP) model that allow for time-dependent travel times; (2) we propose a risk-sensitive criterion that allow for different risk preferences; and (3) we develop a local search algorithm to solve DSOPs with this risk-sensitive criterion. We also empirically show that this approach is applicable on a real-world theme park navigation problem. We anticipate that such user-centric route guidance will be more widely used as users carry smart devices and use mobile applications with increasing intensity.

---

[3]Peak days are Fridays, Sundays and Mondays according to our theme park operator.